\documentclass[10pt,twocolumn,letterpaper]{article}

\usepackage{cvpr}      
\definecolor{cvprblue}{rgb}{0.21,0.49,0.74}
\usepackage[pagebackref,breaklinks,colorlinks,allcolors=cvprblue]{hyperref}

\def\paperID{*****} 
\def\confName{CVPR}
\def\confYear{2025}

\title{Pseudo-Label Enhanced Cascaded Framework: 2nd Technical Report for LSVOS 2025 VOS Track
}

\author{
An Yan\textsuperscript{1} \quad
Leilei Cao\textsuperscript{1} \quad
Feng Lu\textsuperscript{2} \quad
Ran Hong\textsuperscript{1} \quad
Youhai Jiang\textsuperscript{1} \quad
Fengjie Zhu\textsuperscript{1} \\
\textsuperscript{1}TEX AI, Transsion Holdings \quad
\textsuperscript{2}ShanghaiTech University \\
{\tt\small an.yan@transsion.com, leilei.cao@transsion.com}
}
\usepackage{graphicx}
\begin{document}

\maketitle
\begin{abstract}
Complex Video Object Segmentation (VOS) presents significant challenges in accurately segmenting objects across frames, especially in the presence of small and similar targets, frequent occlusions, rapid motion, and complex interactions.
In this report, we present our solution for the LSVOS 2025 VOS Track based on the \textbf{SAM2} framework.
We adopt a pseudo-labeling strategy during training: a trained SAM2 checkpoint is deployed within the \textbf{SAM2Long} framework to generate pseudo labels for the MOSE test set, which are then combined with existing data for further training.
For inference, the SAM2Long framework is employed to obtain our primary segmentation results, while an open-source \textbf{SeC} model runs in parallel to produce complementary predictions. 
A cascaded decision mechanism dynamically integrates outputs from both models, exploiting the temporal stability of SAM2Long and the concept-level robustness of SeC.
Benefiting from pseudo-label training and cascaded multi-model inference, our approach achieves a \textbf{J\&F score of 0.8616} on the MOSE test set --- \textbf{+1.4} points over our SAM2Long baseline --- securing the \textbf{2nd place} in the LSVOS 2025 VOS Track, and demonstrating strong robustness and accuracy in long, complex video segmentation scenarios.
\end{abstract}

\section{Introduction}
\label{sec:intro}

Video Object Segmentation (VOS) is a fundamental task in computer vision, aiming to track and segment target objects throughout a video sequence given only the annotations in the first frame. It has broad applications in autonomous driving, augmented reality, interactive video editing, and automatic data annotation~\cite{perazzi2017davis, xu2018youtube}. In the semi-supervised setting, the model is required to handle arbitrary categories, propagating initial masks to all subsequent frames.

Recent VOS research has been dominated by memory-based approaches. The Space-Time Memory Network (STM)~\cite{oh2019video} stores features and masks from past frames in an external memory and retrieves relevant information for a query frame via pixel-level matching. The Space-Time Correspondence Network (STCN)~\cite{cheng2021rethinking} improves memory reading efficiency by encoding key features without masks and using L2 similarity. XMem~\cite{cheng2022xmem} further introduces a multi-level memory hierarchy—sensory, working, and long-term memories—based on the Atkinson–Shiffrin model, which is particularly effective for long sequences.

Transformer architectures~\cite{vaswani2017attention} have also been integrated into VOS to enhance object-level reasoning. AOT (Associating Objects with Transformers)~\cite{yang2021aot} introduces object queries to maintain consistent identities across frames. Cutie~\cite{cheng2023putting} extends this idea via explicit object memory and bidirectional interactions between pixel-level and object-level features, achieving robustness against occlusions, distractors, and appearance changes.

More recently, advances in foundation segmentation models have brought new capabilities to VOS. Meta released the Segment Anything Model (SAM)~\cite{kirillov2023sam}, a prompt-based foundation model trained on the large-scale SA-1B dataset~\cite{kirillov2023sam}, demonstrating strong zero-shot generalization and competitive accuracy when extended to video data. However, SAM in its original form is not designed for temporal consistency and relies heavily on low-level visual similarity, making it less robust under drastic appearance changes and scene transitions.

SAM2~\cite{ravi2024sam2} extended this paradigm to video object segmentation by incorporating advanced memory mechanisms and leveraging the large-scale SA-V dataset, achieving state-of-the-art results on multiple VOS benchmarks. While SAM2 significantly improves temporal propagation, it still struggles to maintain target identity under severe long-term occlusion or rapid appearance variations.

To better handle long-duration videos, SAM2Long~\cite{ding2024sam2long} further adapts SAM2 with a training-free memory tree, improving temporal stability across extended sequences. Nonetheless, SAM2Long’s reliance on memory-based matching means it can still lose track of objects when they undergo frequent and substantial appearance changes.

To address the limitations of purely memory-based matching, concept-driven approaches such as SeC~\cite{zhang2025sec} leverage large vision-language models to progressively construct object-level semantic representations, enabling robust segmentation under challenging conditions with drastic appearance changes or shot transitions. While SeC excels at maintaining target identity across diverse scenes, its reliance on high-level semantic abstraction can lead to less accurate boundary localization for small or fine-scale objects, especially in crowded scenarios. This makes SeC highly complementary to the temporal stability and fine-grained localization capability of SAM2Long, motivating our combined framework for complex long-video segmentation.

In terms of benchmarks, DAVIS~\cite{perazzi2017davis} and YouTube-VOS~\cite{xu2018youtube} have driven early progress in VOS, yet they mostly feature salient objects in clean backgrounds. More recent datasets have expanded scale and diversity: SA-1B~\cite{kirillov2023sam} enables large-scale training of foundation segmentation models; SA-V~\cite{ravi2024sam2} provides 51k annotated videos for video segmentation; LSVOS v2~\cite{hong2024lvosv2} focuses on long-duration videos with persistent object interactions. The MOSE dataset~\cite{ding2023mose}, adopted in the LSVOS 2025 VOS Track, presents additional challenges: small and similar objects, frequent occlusions, abrupt appearance changes, and disappearance–reappearance events. These conditions exacerbate ID switches, drifting, and incomplete masks for existing approaches.

\begin{figure}[h]
  \centering
  \includegraphics[width=\linewidth]{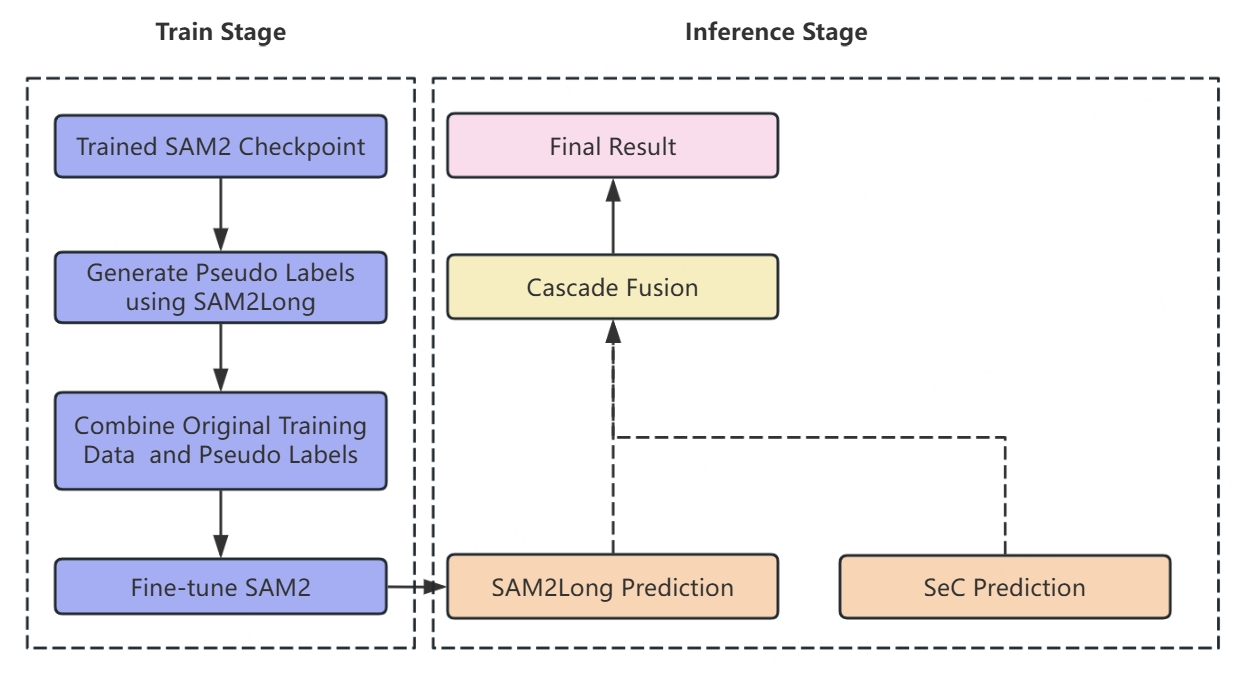} 
  \caption{Overview of our method: pseudo-label enhanced SAM2Long training and cascaded inference with SeC.}
  \label{fig:flowchart}
\end{figure}

Motivated by the strengths and limitations of existing methods, we propose a solution built upon the strong segmentation capability of the SAM2 framework. Notably, in the CVPR2022 Large-scale Video Object Segmentation Challenge (Track 3: Referring VOS), Cao et al.~\cite{cao2022rvos} achieved second place by enhancing the ReferFormer baseline with cyclical learning rates, semi-supervised training via pseudo labels, and test-time augmentation. Their success demonstrates the effectiveness of leveraging unlabeled data through pseudo labeling in competitive settings, which partially inspires our pseudo-labeling design.As shown in Fig.~\ref{fig:flowchart}, during training, we adopt a pseudo-labeling strategy: a trained SAM2 checkpoint is deployed within the SAM2Long framework to generate pseudo labels for the MOSE test set, which are then combined with existing annotated data for retraining to reduce domain gap and improve robustness. For inference, the SAM2Long framework generates the primary segmentation results, while a SeC model is executed in parallel to produce complementary predictions. The two results are fused through a cascaded combination, effectively integrating the strengths of both models.

With this hybrid strategy, our method achieves a \textbf{J\&F score of 0.8616} on the MOSE test set and ranks \textbf{second place} overall, demonstrating strong robustness and accuracy in long and complex video object segmentation scenarios.

\section{LSVOS Tracks and Datasets}

The Large-Scale Video Object Segmentation (LSVOS) Challenge 2025 consists of three independent competition tracks, each associated with a specific dataset designed to evaluate different aspects of video object segmentation under challenging real-world conditions. Table~\ref{tab:lsvos_datasets} presents an overview of the datasets used in each track.

\begin{table}[ht]
\centering
\caption{Statistics of the datasets used in the three tracks of the LSVOS 2025 Challenge.}
\label{tab:lsvos_datasets}
\begin{tabular}{l|c|c|c|c}
\hline
\textbf{Dataset} & \textbf{Videos} & \textbf{Categories} & \textbf{Objects} & \textbf{Masks} \\ \hline
MOSEv1  & 2,149 & 36  & 5,200 & 431,725 \\
MOSEv2  & 5,024 & 200 & 10,074 & 701,976 \\
MeViS   & 2,006 & --  & 8,171 & 443,000 \\
\hline
\end{tabular}
\end{table}

\subsection{Track 1: MOSEv2 -- Complex Video Object Segmentation}
The second track adopts MOSEv2~\cite{ding2025mosev2}, an enhanced and expanded version of MOSEv1 with significantly greater scale and complexity. MOSEv2 includes 5,024 videos spanning 200 categories, with 10,074 annotated objects and over 701,976 high-quality masks. This dataset not only preserves the challenges from MOSEv1 (disappearance--reappearance, small objects, occlusion, and crowding) but also introduces new real-world difficulties such as adverse weather (rain, snow, fog), low-light scenes (nighttime, underwater), multi-shot sequences, camouflaged objects, non-physical targets (shadows, reflections), and knowledge-dependent scenarios. MOSEv2 has a Disappearance Rate of 61.8\%, a Reappearance Rate of 50.3\%, and an average of 13.6 similar distractors per target. Furthermore, 50.2\% of instances are small objects (mask area $<$1\% of frame), posing substantial challenges for fine-grained segmentation across a variety of extreme conditions.

\subsection{Track 2: MOSEv1 -- Video Object Segmentation}
The first track is based on the MOSEv1 dataset~\cite{ding2023mose}, targeting semi-supervised video object segmentation (VOS) in complex scenes. MOSEv1 contains 2,149 high-resolution videos with 5,200 annotated objects from 36 categories, totaling 431,725 high-quality segmentation masks. Compared to earlier benchmarks such as DAVIS and YouTube-VOS, MOSEv1 introduces significantly more challenging scenarios, including frequent object disappearance and reappearance, small or inconspicuous targets, heavy occlusion, and crowded environments. These factors greatly increase the difficulty of maintaining temporal consistency and segmentation accuracy, making MOSEv1 a valuable benchmark for evaluating robustness in realistic scenes.

\subsection{Track 3: MeViS -- Referring Video Object Segmentation}
The third track is built on the MeViS dataset~\cite{ding2023mevis,ding2025mevis}, a large-scale benchmark for Referring Video Object Segmentation (RVOS) driven primarily by motion expressions. Unlike existing RVOS datasets that rely heavily on static attributes in language descriptions, MeViS focuses on motion cues requiring temporal reasoning across the whole video. The dataset contains 2,006 videos with 8,171 annotated objects and 28,570 natural language expressions, resulting in 443,000 segmentation masks. Each expression may refer to one or multiple targets (multi-object expressions), increasing the task complexity. MeViS features scenes with multiple visually similar objects, requiring algorithms to distinguish targets based on motion patterns rather than appearance salience or category labels. This track evaluates the capability to integrate language understanding with temporal motion analysis for accurate segmentation in complex, dynamic environments.

\section{Method}
\label{sec:method}

Our solution for the LSVOS 2025 VOS Track is built upon the SAM2 framework, enhanced by pseudo-label based domain adaptation and cascaded inference with the SeC model. Fig.~\ref{fig:flowchart} illustrates the overall pipeline, which consists of four main components: (1) a SAM2Long-based baseline, (2) pseudo-label generation, (3) retraining with pseudo labels, and (4) cascaded multi-model inference.

\subsection{Baseline: SAM2 and SAM2Long}

We adopt the Segment Anything Model 2 (SAM2) as our core segmentation backbone due to its strong generalization capability across diverse video object segmentation scenarios. Specifically, we employ the ViT-L variant of SAM2 as the base architecture. To better handle long video sequences, we leverage the SAM2Long framework~\cite{ding2023mose}, which extends SAM2's temporal modeling capacity via memory propagation mechanisms, making it well-suited for datasets such as MOSE.

In our implementation, the ViT-L variant of SAM2 is first fine-tuned on the target-domain data. The resulting checkpoint serves as the foundation for subsequent enhancements.

\subsection{Pseudo-label Generation}

To bridge the domain gap between the available training data and the MOSE test distribution, we adopt a pseudo-labeling strategy. Specifically, the fine-tuned SAM2 (ViT-L) checkpoint is used within the SAM2Long framework to generate segmentation masks for each frame of the unlabeled MOSE test set. No additional post-processing or low-confidence filtering is applied, ensuring the complete output distribution is preserved. These generated pseudo labels are then combined with the original MOSE training data to produce an augmented dataset.

\subsection{Retraining with Pseudo Labels}

The augmented dataset (original training data + MOSE pseudo-labeled test data) is used to further fine-tune the SAM2 model. All data augmentation strategies from the official SAM2 fine-tuning protocol are retained, including random scaling, random cropping, horizontal flipping, and color jittering. The model is trained with a batch size of 1 for 45 epochs, following the same optimizer, learning rate schedule, and loss functions as in the original SAM2 fine-tuning setup. This retraining significantly improves the model's robustness in the MOSE test domain.

\subsection{Performance on MOSE Validation Set}

We evaluate each stage of our pipeline on the \textbf{MOSE validation set} to quantify the improvements from fine-tuning, pseudo-labeling, and SAM2Long. As shown in Table~\ref{tab:baseline-valid}, the open-source SAM2 (ViT-L) achieves a J\&F score of \textbf{0.7428}. Fine-tuning on the MOSE training set improves this to \textbf{0.7624}, while incorporating pseudo-labeled data further increases it to \textbf{0.7700}. Initializing SAM2Long with the latter checkpoint boosts performance to \textbf{0.7788}, highlighting its enhanced temporal consistency.

\begin{table}[h]
\centering
\caption{Performance improvements on the MOSE validation set from fine-tuning, pseudo-labeling, and SAM2Long.}
\label{tab:baseline-valid}
\begin{tabular}{p{5cm}|c} 
\hline
\textbf{Model \& Training Strategy} & \textbf{J\&F (MOSE valid)} \\
\hline
SAM2  & 0.7428 \\
SAM2 (fine-tuned on MOSE train) & 0.7624 \\
SAM2 (fine-tuned + pseudo labels) & 0.7700 \\
SAM2Long  & \textbf{0.7788} \\
\hline
\end{tabular}
\end{table}

\subsection{Cascade Inference with SeC}

Our default segmentation backbone for inference is \textbf{SAM2Long}, initialized with the SAM2 (ViT-L) checkpoint fine-tuned on MOSE train data and augmented with pseudo-labeled MOSE test data. While SAM2Long provides strong performance for long-term tracking, it may fail in scenarios with frequent and drastic appearance variations. To address this, we integrate the Segment Concept (SeC) model~\cite{zhang2025sec} as a secondary inference branch, and design a cascade decision strategy to dynamically select the more reliable predictions at the video level.

\begin{figure}[h]
\centering
\includegraphics[width=\linewidth]{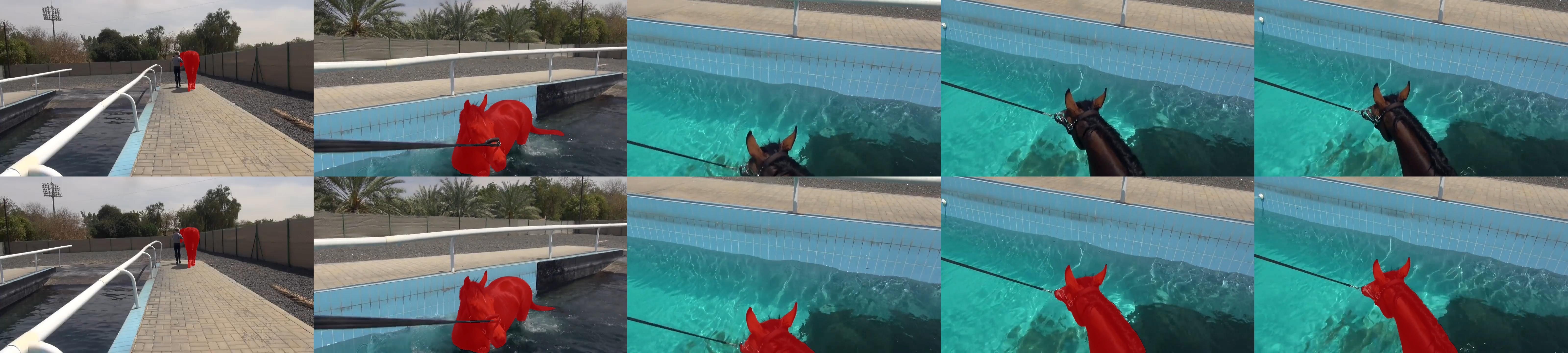}
\caption{\textbf{Example of Miss Tracking.} Top row: SAM2Long predictions; bottom row: SeC predictions. SAM2Long fails to output valid masks in some frames, whereas SeC successfully segments the target.}
\label{fig:miss_tracking_example}
\end{figure}

We define model disagreement as frames where the IoU between SAM2Long and SeC predictions is $\leq 0.1$, and categorize it into two cases with \textbf{priority}:

\textbf{1) Miss Tracking (Higher Priority):}  
If one model outputs a valid mask while the other does not, the frame is counted as a miss tracking case.  
If the total number of such frames (non-consecutive, accumulated across the video) exceeds $10$, we directly select the predictions from the model with valid masks for the entire video, without applying the wrong tracking criterion.  
An example of such a case is shown in Figure~\ref{fig:miss_tracking_example}, where SAM2Long fails to output a mask for several frames, while SeC predicts valid masks for the same frames.

\begin{figure}[h]
\centering
\includegraphics[width=\linewidth]{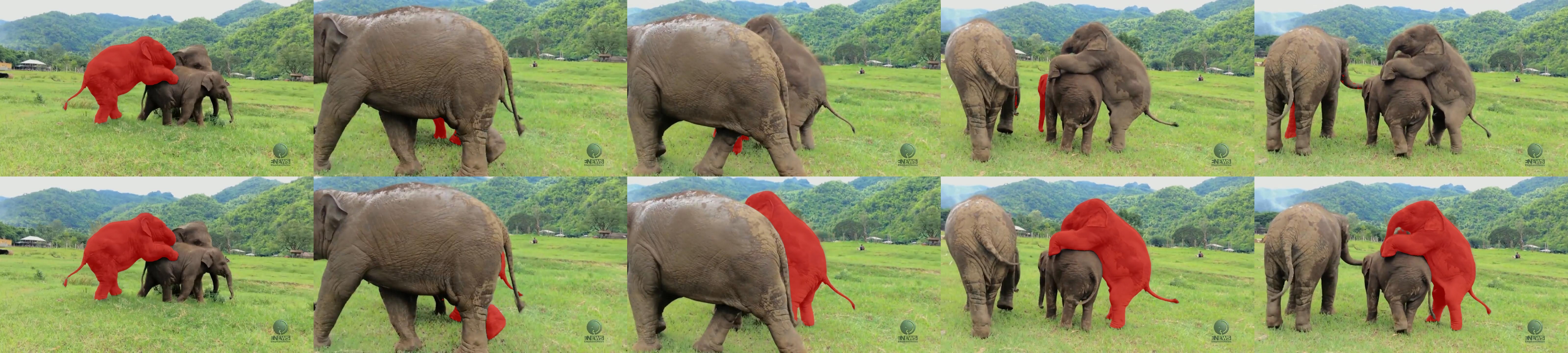}
\caption{\textbf{Example of Wrong Tracking.} Top row: SAM2Long predictions; bottom row: SeC predictions. In these frames, both models produce masks but with IoU $\leq 0.1$. The noisier masks (more fragmented contours) indicate worse tracking quality in those frames.}
\label{fig:wrong_tracking_example}
\end{figure}

\textbf{2) Wrong Tracking:}  
We first count the number of frames in a video where both models output valid masks but the IoU between them is still $\leq 0.1$.  
If this number exceeds 10 frames, we further evaluate mask noise on these disagreement frames by counting the total number of mask contours for each object ID.  
Frames with a contour count greater than 6 are considered high-noise frames.  
The model with fewer high-noise frames across these disagreement frames is selected as the final output for the entire video.  
An example of Wrong Tracking is shown in Figure~\ref{fig:wrong_tracking_example}, where both models produce masks but with very low IoU, and one model generates noisier masks than the other.

This cascade mechanism exploits SeC's adaptability in handling severe appearance changes while maintaining SAM2Long's stability in consistent scenarios.

On the \textbf{MOSE test set}, as shown in Table~\ref{tab:main_results}, SeC alone achieves a J\&F score of \textbf{0.8246}, SAM2Long (initialized with the fine-tuned + pseudo-labeled SAM2 checkpoint) scores \textbf{0.8475}, and our cascade inference further boosts performance to \textbf{0.8616}.

\begin{table}[h]
\centering
\caption{Comparison of different model variants on the MOSE 2025 VOS Track (test set). All scores are $\mathcal{J\&F}$.}
\label{tab:main_results}
\begin{tabular}{l|c}
\hline
Method & $\mathcal{J\&F}$  \\
\hline
SAM2Long (with pseudo-label training) & 0.8475 \\
SeC (open-source model) & 0.8246 \\
Ours (Cascade: SAM2Long + SeC) & \textbf{0.8616} \\
\hline
\end{tabular}
\end{table}

This cascaded inference mechanism enables our system to dynamically exploit the strengths of SeC in challenging appearance-changing scenes while relying on SAM2Long for stable scenarios.

\section{Experiments}
\label{sec:experiments}

\subsection{Dataset and Evaluation Protocol}
We evaluate our approach on the MOSE test set, the benchmark dataset used in the LSVOS 2025 VOS Track~\cite{ding2023mose}.  
The dataset contains diverse and challenging real-world videos, featuring frequent occlusions, small and similar objects, fast motion, and object disappearance–reappearance events.

All evaluations are conducted under the semi-supervised setting, where the ground-truth mask of the first frame is provided for each target object.  
Following the official evaluation protocol, we report region similarity $\mathcal{J}$, contour accuracy $\mathcal{F}$, and their average score $\mathcal{J\&F}$.

\subsection{Implementation Details}
Our SAM2 baseline is fine-tuned with pseudo labels generated from the MOSE test set, as described in Section~\ref{sec:method}.  
The final system integrates SAM2Long and the SeC model through a cascaded decision scheme.  
All experiments are performed on a single NVIDIA H200 GPU using frames resized to $1024 \times 1024$.  

\begin{table*}[t]
\centering
\caption{Ablation study of pseudo-label training, SAM2Long adaptation, and cascaded inference on MOSE validation and test sets.}
\label{tab:ablation_all}
\begin{tabular}{l|c|c}
\hline
\textbf{Method / Optimization Stage} & $\mathcal{J\&F}$ (valid) & $\mathcal{J\&F}$ (test) \\
\hline
SAM2 (ViT-L, fine-tuned on MOSE train) & 0.7624 & - \\
+ Pseudo-label training (SAM2) & 0.7700 & - \\
SAM2Long (initialized with above checkpoint) & 0.7788 & - \\
SAM2Long (with pseudo-label training) & - & 0.8475 \\
SeC (open-source) & - & 0.8246 \\
\textbf{+ Cascade inference (SAM2Long + SeC)} & - & \textbf{0.8616} \\
\hline
\end{tabular}
\end{table*}

\subsection{Main Results on MOSE Test Set}
Table~\ref{tab:main_results} presents the comparison between our approach and individual model variants on the MOSE test set.  
The pseudo-label training significantly improves SAM2Long from its baseline performance, while the SeC model, despite being an open-sourced checkpoint, exhibits strong performance in scenarios with frequent appearance variations.  
By combining the strengths of both models through our cascaded inference strategy, the final system achieves a $\mathcal{J\&F}$ of 0.8616, ranking \textbf{2nd place} in the official track leaderboard.

\subsection{Ablation Study}
We further analyze the contribution of the cascaded inference mechanism, as shown in Table~\ref{tab:ablation_all}.  
Starting from SAM2Long with pseudo-label training ($\mathcal{J\&F} = 0.8475$), introducing the SeC branch within our cascaded selection scheme pushes performance to 0.8616, confirming the complementarity between SAM2Long's temporal stability and SeC's robustness to severe appearance changes.

\subsection{Qualitative Results}
\begin{figure}[h]
    \centering
    \includegraphics[width=\linewidth]{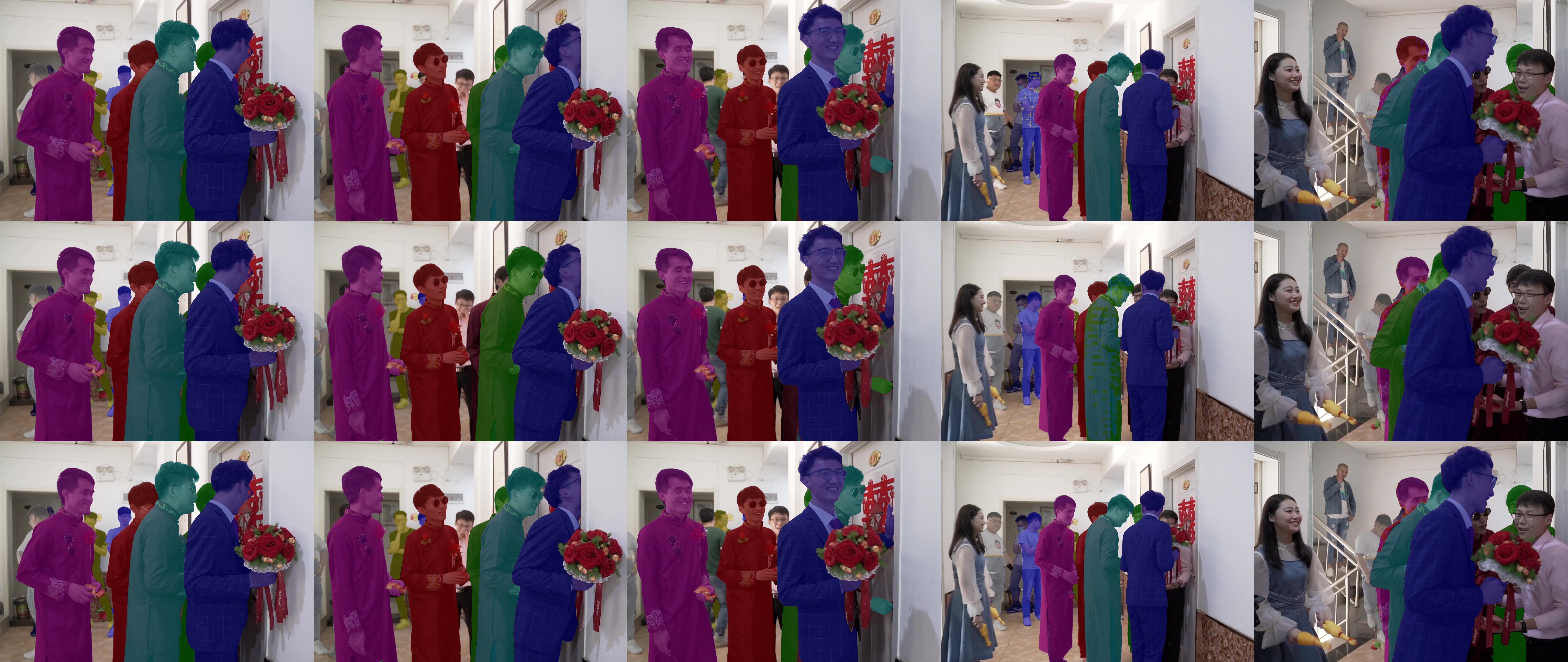}
    \caption{Qualitative comparison among SAM2Long (top row), SeC (middle row), and our cascaded output (bottom row). 
        SeC yields finer segmentation boundaries in challenging scenarios with large appearance changes or occlusions. 
        Our cascaded output integrates the strengths of both methods, resulting in more accurate and consistent predictions.}
    \label{fig:qualitative}
\end{figure}

As shown in Figure \ref{fig:qualitative} presents qualitative comparisons among SAM2Long, SeC, and our final cascaded output.
In challenging scenarios with substantial appearance variations or severe occlusions, SeC demonstrates more precise segmentation boundaries.
These refined boundaries are effectively integrated into the final predictions through our decision mechanism, leading to improved overall segmentation quality.

\section{Conclusion}
In this report, we presented our solution for the LSVOS 2025 VOS Track, which combines the strong segmentation capability of SAM2Long with the concept-level reasoning power of SeC through a cascaded inference strategy. A pseudo-labeling scheme was employed to adapt SAM2Long to the MOSE data domain, significantly improving its baseline performance. The cascaded decision mechanism dynamically integrates outputs from SAM2Long and SeC, leveraging their respective strengths to handle both stable and complex, appearance-changing scenarios in long videos. Our final system achieves a $\mathcal{J\&F}$ score of 0.8616 on the MOSE test set, ranking \textbf{2nd place} overall in the official competition. In the future, we plan to further enhance concept-level reasoning integration and explore more advanced pseudo-label refinement strategies to boost robustness in extreme video segmentation conditions.

{
    \small
    \bibliographystyle{ieeenat_fullname}
    \bibliography{main}
}


\end{document}